\documentclass[preprint,12pt]{elsarticle}

\usepackage{amssymb, adjustbox}

\journal{Journal of Informetrics}

\begin{document}

\begin{frontmatter}



\title{Combining keyphrase extraction and lexical diversity to characterize ideas in publication titles}


\author[inst1]{James Powell}
\ead{jepowell@lanl.gov}

\affiliation[inst1]{organization={Los Alamos National Laboratory},
            addressline={PO Box 1663}, 
            city={Los Alamos},
            postcode={87545}, 
            state={New Mexico},
            country={USA}}

\author[inst1]{Martin Klein}

\author[inst1]{Lyudmila Balakireva}

\begin{abstract}
Beyond bibliometrics, there is interest in characterizing the evolution of the number of ideas in scientific papers. A common approach for investigating this involves analyzing the titles of publications to detect vocabulary changes over time. With the notion that phrases, or more specifically keyphrases, represent concepts, lexical diversity metrics are applied to phrased versions of the titles. Thus changes in lexical diversity are treated as indicators of shifts, and possibly expansion, of research. Therefore, optimizing detection of keyphrases is an important aspect of this process. Rather than just one, we propose to use multiple phrase detection models with the goal to produce a more comprehensive set of keyphrases from the source corpora. Another potential advantage to this approach is that the union and difference of these sets may provide automated techniques for identifying and omitting  non-specific phrases. We compare the performance of several phrase detection models, analyze the keyphrase sets output of each, and calculate lexical diversity of corpora variants incorporating keyphrases from each model, using four common lexical diversity metrics.
\end{abstract}

\begin{keyword}
natural language processing \sep keyphrase detection \sep lexical diversity
\end{keyword}

\end{frontmatter}


\section{Introduction}
Research into determining the cognitive extent \cite{milojevic2015quantifying} of scholarly output is a new field. It is largely motivated by the goal of quantifying the growth of ideas in scientific publications over time. It blends bibliometric practices with Natural Language Processing (NLP) to break down the barrier between counting scholarly output and determining what it is about. The goal is to establish systematic techniques for characterizing the introduction and evolution of new ideas and research topics. This necessitates identifying the best proxies for distinct ideas in publications, and finding reasonable ways to characterize them over time. The results of such analysis would enhance understanding of research outputs, and could compliment traditional bibliometrics and their role in planning, and funding efforts that affect future research. 

Our goal was to expand upon prior work in this area by combining multiple keyphrase detection approaches with multiple lexical diversity metrics so as to evaluate various combinations of each. The  work of Milojevic et al (\cite{milojevic2015quantifying}) and Berube et al \cite{berube2018words} has demonstrated that lexical diversity can provide valid insights into the cognitive extent of a corpus, even if the corpus is limited to publication titles.  To ensure our work was comparable to prior work, we used only publication titles to form the basis of our test corpus. These titles represented publications spanning a decade. We identified and applied several distinct automated techniques for extracting keyphrases from titles. We then applied multiple diversity metrics to the output to determine if different metrics yield comparable values for the diversity of a collection of titles in a given year. 

\section{Prior Work}
\subsection{Detecting keyphrases}
Keyphrase extraction has been explored for many natural language processing tasks, most often utilizing full text when it is available  \cite{nguyen2007keyphrase}, \cite{kim2001corpus}, and \cite{turney2000learning}. Relatively few projects have attempted to extract keyphrases from short text or publication titles. Keyphrase Extraction and Ranking by Topic (KERT) aims to generate topical keyphrases from short texts, arguing that "content-representative titles are cleaner and more efficient to deal with than entire documents. \cite{li2010semi} also explores keyphrase detection in titles, noting that "the title has a similar role to the key phrases" and arguing that when a keyphrase occurs in a title, it ought to be ranked higher.  \cite{bhowmik2008keyword} proposed using a simple neural network to extract keyphrases from titles and abstracts, and they compared their results with author provided keyphrases. Another often mined collection of short texts are available on Twitter and they pose similar challenges for keyphrase extraction as document titles. Most of this work has been concerned with summarizing tweets via keyphrase extraction (\cite{zhao2011topical}, \cite{zhang2016keyphrase}, and \cite{bellaachia2012ne}).  

\subsection{Lexical diversity and scientific literature}
Lexical diversity has been applied to a range of language evaluation use cases, including language acquisition, language comprehension, assessment of language disorders and language impairment, and generally to measure an individual's reading, speaking, or  writing skills. It is a measure of the richness of the vocabulary used or comprehended by an individual. General guidance from McCarthy et al recommends that researchers “consider using [multiple metrics] in their studies, rather than any single index, noting that lexical diversity can be assessed in many ways and each approach may be informative as to the construct under investigation.” \cite{mccarthy2010mtld} In recent years, we know of only two studies that have used lexical diversity to characterize the concepts and ideas in scientific publication titles. \cite{milojevic2015quantifying} explored lexical diversity as a means to "quantify the extents of cognitive domains of different bodies of scientific literature" using publication titles he found that " periods of cognitive growth do not necessarily coincide with the trends in publication volume." \cite{berube2018words} developed an approach to "assess the evolution of lexical diversity in scholarly titles using a new indicator based on Zipfian frequency-rank distribution tail fits." They found that they could measure lexical concentration of titles as well as predict the leveling off of vocabulary expansion in a given field. 

\section{Research Goals}
The objective of this paper is to investigate whether and how lexical diversity changes depending on how keyphrases are identified in a corpus. Our hypothesis is that keyphrases represent the best proxies for research ideas and concepts, and that measurements of keyphrase diversity would be an optimal way to evaluate cognitive expansion in a corpus over time. We believe that by using multiple keyphrase detection techniques combined with multiple lexical diversity metrics, we could simultaneously improve the comprehensiveness of concept detection, and validate (or invalidate) lexical diversity as a way to measure change and growth of ideas. To evaluate this hypothesis we performed the following tasks 
\begin{enumerate}
    \item identify and apply several distinct approaches for detecting keyphrases in a title corpus. 
    \item Create a version of the title corpus using each phrase model and  evaluate each using four different lexical diversity metrics. 
    \item Compare the lexical diversity scores. 
    \item Characterize the intersection among the phrase detection algorithms, as well as the collection of “leftover” phrases, that is, the difference between the various phrase sets, when comparing phrase detection approaches. 
\end{enumerate}

Items 3 and 4 should provide evidence for or against the benefit of employing multiple keyphrase detection approaches. 

\subsection{Data and NLP Preprocessing Pipeline}
Our test corpora consisted of the titles of publicly available reports and papers in our local library repository. Our organization employs a pre-release review, which means that most publications intended for external publication are submitted by the primary author and stored in a local database. Only items that are not flagged as drafts, and are not embargoed, are included in our data. We used publication titles spanning 2012 to 2020 (see table  ~\ref{tabsummary}). Since there is sometimes a lag for submissions, the number of 2020 articles may be lower due to items not yet submitted or made public. We employed several means for identifying lexical units that reflect distinct concepts, or keyphrases, including statistical techniques as well as part of speech tagging and extraction. These are described in the methods section below. 

Basic text normalization was employed including lower case, tokenization via punctuation, and a very small stopwords list was also used, consisting of 'a, an, the.' Since many phrases may contain words that are usually considered stopwords, we elected not to use a full stopword list. By removing these three words, we vastly reduced the number of article + noun phrase constructs identified by the part of speech model. Since we were interested in keyphrases and not all words in titles, this approach allowed us to primarily accumulate more complex n-gram phrases.

\section{Methods}

\subsection{Keyphrase detection approaches}
As in \cite{berube2018words}, we are interested in extracting and analyzing keyphrases, rather than just keywords or all words. We focused on fast, simple algorithms for keyphrase extraction, that require minimal training, are unsupervised, or were based on available off-the-shelf models. We extract keyphrases from titles, reducing each title to its constituent phrases. This results in a distinct corpus variant per keyphrase detection algorithm. Then we measure the lexical diversity of these variants of the original title corpus using four different lexical diversity metrics, one due to its intuitive nature, and three others which have been empirically demonstrated to produce diversity scores that are less susceptible to  the variable text length problem that plague some lexical diversity metrics. 

We selected three techniques for detecting and extracting keyphrases in the title corpora: 
\begin{itemize}
\item A pre-trained phrase model trained only on the aggregation of all titles spanning 2012-2020 (the title phraser), 
\item A pre-trained phrase model trained on several decades of (our organization) reports (the full-text phraser),
\item Noun phrase detection based on part of speech detection supported by a large English language model implemented in the SPACY NLP library. 
\end{itemize}

The title and full-text phrases use the same training algorithm with different size corpora: Pointwise Mutual Information (PMI) \cite{bouma2009normalized}. The objective of PMI is to establish the likelihood that two words co-occur as a single lexical unit, as reflected by equation ~\ref{pmi}

\begin{equation}
 \label{pmi}
 PMI(x,y) = \log_2 \frac{P(x,y)} {P(x)P(y)}
\end{equation}

\noindent where a PMI score of 0 indicates the two variables (words) are independent, while a positive value indicates the two terms occur together more than would be expected if they were unrelated. In practice, one of the PMI implementations in the gensim \cite{rehurek2011gensim} NLP library takes into account the overall frequency of individual words and the size of the training corpus (equation ~\ref{bigram}):

\begin{equation}
    \label{bigram}
    \frac {(bigram\_count - min\_count)(len\_vocab)} {(worda\_count)(wordb\_count)}
\end{equation}

\noindent In this case, $bigram\_count$ is the PMI score as determined by the prior equation. 
This discounts a high PMI score that is the result of one or both of the terms being broadly common in the corpus. Implementations of PMI analysis for phrase detection in text may be trained on any text corpus, but it is more common to use a corpus of text that is identical to, or a relevant superset of the text upon which phrase detection is to be performed. PMI has been compared to other techniques for NLP tasks and in some cases found to approach human-level judgement for some tasks \cite{recchia2009more}.

SPACY \cite{spacy2}utilizes part of speech detection in order to identify noun phrases in text sequences. Noun phrases have been explored in the past for improving access to content in digital libraries \cite{tolle2000comparing}. Part of speech detection is best performed over the original text rather than text which has been modified by stopword elimination and lemmatization. As mentioned above we do remove three articles from the corpus, but otherwise the original titles are presented as input with punctuation introduced at the end of each title (a period) to provide SPACY with the equivalent of sentences rather than a continuous stream of text. We used the large SPACY english language model (en\_core\_web\_lg).

We applied each title phrase extraction model to each year of the title corpus. These titles were reduced to only the phrases detected by the phrase extraction model. This resulted in three modified versions of the title corpus, one per phrase extraction model. For comparison, we also used unprocessed versions of the corpora in their original form, with no keyphrase detection applied (we refer to the unmodified title corpora  as the “all words” version). Table ~\ref{tabphrasing} illustrates the conversion process from the original sentence to various phrased forms of the sentence.

\begin{table}[h]
\caption{Original sentences, phrased variants}
\label{tabphrasing}
\begin{tabular}
{|p{0.25\columnwidth}|p{0.75\columnwidth}|}
\hline
\textbf{Sentence Form}&\textbf{After phrase detection}\\
\hline
Original sentence&Genetic diversity within Clostridium botulinum serotypes, botulinum neurotoxin gene clusters and toxin subtypes.\\
\hline
Title phraser&'diversity\_within', 'botulinum\_neurotoxin'\\
\hline
Full-text phraser&'genetic\_diversity', 'clostridium\_botulinum', 'botulinum\_neurotoxin',  'gene\_clusters', 'toxin\_subtypes'\\
\hline
Noun phrases&'genetic\_diversity', 'clostridium\_botulinum\_serotypes',  'botulinum\_neurotoxin\_gene\_clusters', 'toxin\_subtypes'\\
\hline

    \end{tabular}
\end{table}

\subsection{Lexical diversity metrics}
Lexical diversity is broadly defined as a measure of the variability and richness of a given text. The simplest way to compute this is to count the number of unique words in a text and divide this value by the total word count. This metric is called the type-token ratio (TTR). For example, if a document has a total of 24 words and 15 of which are unique, then the TTR score is 0.625 (15/24). If the document contains 19 unique words, the TTR score would be 0.79. So the higher the TTR, the greater the lexical diversity. It may not be immediately apparent, but this approach is sensitive to text length. This is due to the fact that the total number of tokens will always grow more quickly than the unique token count, so the TTR score can only decrease as the length of text increases. This makes TTR an unstable metric for comparing texts, if there's a significant difference in the length of the texts being compared. Since TTR is problematic for many lexical diversity use cases, a number of alternate metrics have been proposed which aim to neutralize the effect of variable text length. Many of these metrics employ statistical sampling, or normalization and scaling.

Two examples that use statistical sampling are hypergeometric density (HD-D) and Measure of Textual Lexical Diversity (MTLD). \cite{treffers2013measuring}. HD-D starts with a list of all of the unique words in a corpus. A score is calculated for each word based on the observed probability of finding it in a random sample of 42 words from a text. The final HD-D score for a corpus is the sum of the probabilities for all distinct token types. 

The Measure of Textual Lexical Diversity (MTLD) uses the TTR described above in conjunction with a sampling technique. The text is split into segments and the segment sizes are increased or decreased until a TTR score of .72 is achieved for a given segment. The final MTLD score is the mean text length for of all of these segments. The HD-D sample size (42) and the target score for MTLD (.72) were arrived at empirically by the researchers who developed these respective metrics. 

An example lexical diversity metric that uses normalization is the Maas index \cite{treffers2013measuring}. The Maas index is normalized and scaled version of the TTR, which is represented by equation  ~\ref{maas}: 

\begin{equation}
    \label{maas}
 \frac{ \log N - \log V(N)} {\log _{2}(N)}
\end{equation}

\noindent where N is the number of tokens and V is the number of distinct token types. 
 
Generally speaking, HD-D and MTLD scores will increase as lexical diversity increases.  However, the Maas index will decrease as lexical diversity increases.

\begin{table*}[h]
\caption{Summary of corpora and corpora variants: change over time}
\label{tabsummary}
\begin{tabular}
{|p{0.10\columnwidth}|p{0.15\columnwidth}|p{0.15\columnwidth}|p{0.15\columnwidth}|p{0.15\columnwidth}|p{0.15\columnwidth}|p{0.15\columnwidth}|}
\hline
\multicolumn{3}{|c|}{\textbf{Corpus statistics}} & \multicolumn{4}{c|}{\textbf{Unique Tokens $\Delta$  \%}}\\
\hline
\textbf{Year} &\textbf{Document count}&\textbf{$\mu$ title length}&\textbf{All words}&\textbf{Title phraser phrases}&\textbf{Full-text phraser phrases}&\textbf{Noun phrases}\\
\hline
2012&7093&7.08&&&&\\
\hline
2013&9608&7.03&14.32&14.09&20.35&21.73\\
\hline
2014&9713&7.02&0.8&3.32&-0.52&-1.3\\
\hline
2015&9840&6.86&1.59&-1.55&-0.96&0.58\\
\hline
2016&9715&6.93&-1.81&4.77&-1.93&-2.13\\
\hline
2017&11488&7.2&12.93&11.37&18.46&17.72\\
\hline
2018&11875&7&-0.35&-2.76&-3.96&1.08\\
\hline
2019&12747&6.99&4.76&-0.72&-1.054&5.67\\
\hline
2020&10477&6.94&-4.76&-9.49&-25.9&-12.13\\
\hline
\end{tabular}
\end{table*}

\section{Results and Analysis}
\subsection{Lexical diversity of variants of the title corpora}
\begin{figure}[h]
  \centering
  \includegraphics[width=\linewidth]{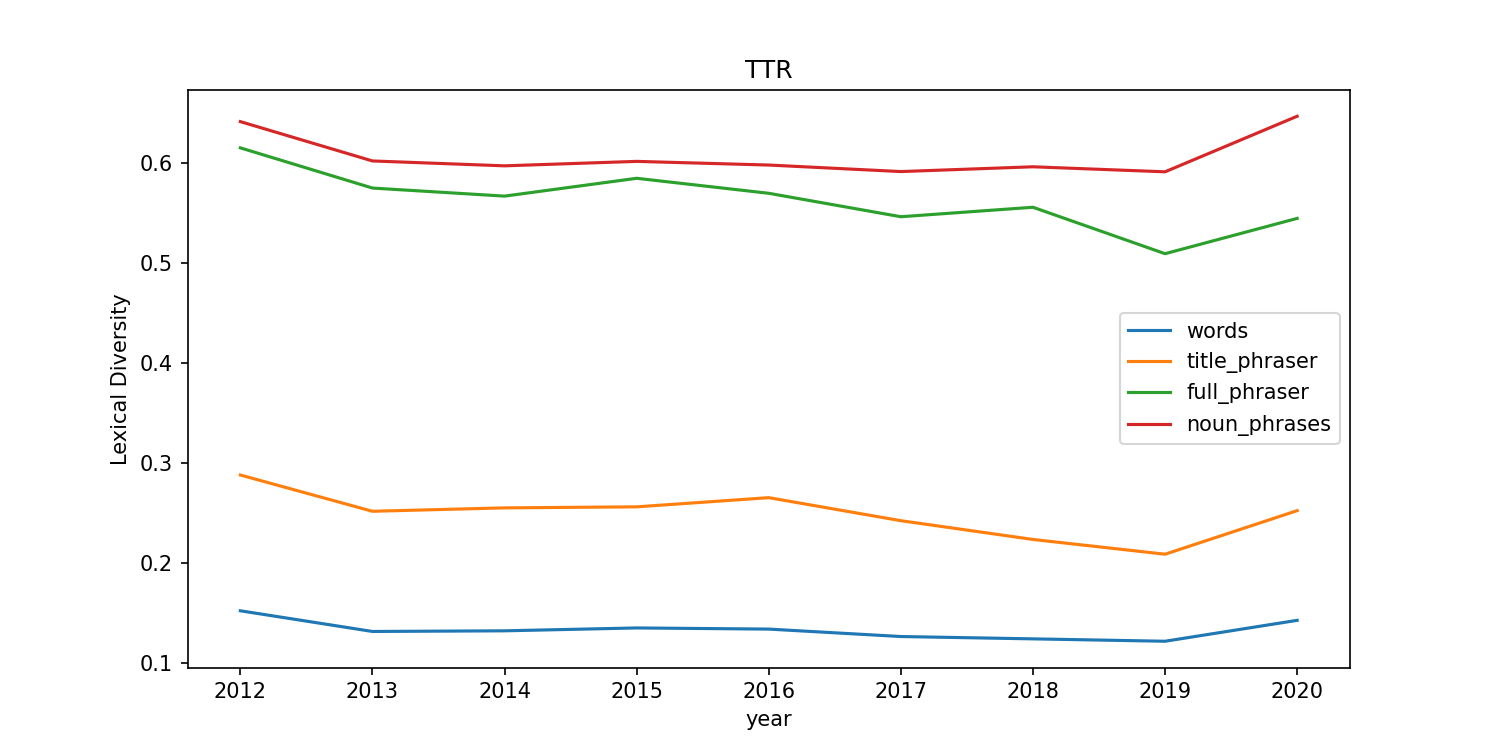}
  \caption{Type-Token Ratio scores for for corpora variants over time}
  \label{fig:ttr}
\end{figure}

Table ~\ref{tabsummary} contains information about each of the four corpora, including year, document count, mean title length, and the percent change for unique lexical unit types over time. Percentage change is in relation to the previous year, so for example in 2013, there was a 14.32\% increase in the word count, a 14.09\% increase in detected title phrases, etc. We analyze each version of the corpora using the four lexical diversity metrics described above (TTR, MTLD, Maas, HD-D) applying them to each of the four versions of the title corpus. This is repeated for each year of report titles. The MTLD index (figure ~\ref{fig:mtld}) for (our organization) reports spanning 2012-2020 exhibits the most significant and persistent shift for the various corpora, starting in 2018. MTLD scores were also highly correlated across all versions of the title corpus, including variants with noun phrases, phrases from the full-text model, and phrase from the  titles only model. For the other three metrics, TTR (figure ~\ref{fig:ttr} had four highly correlated relationships out of six. There were only one pair of title representations that were highly correlated for Maas (Keywords and Noun phrases). Similarly, only one relationship was highly correlated per HD-D scores (Full-text phraser and Noun phrases).

\begin{figure}[h]
  \centering
  \includegraphics[width=\linewidth]{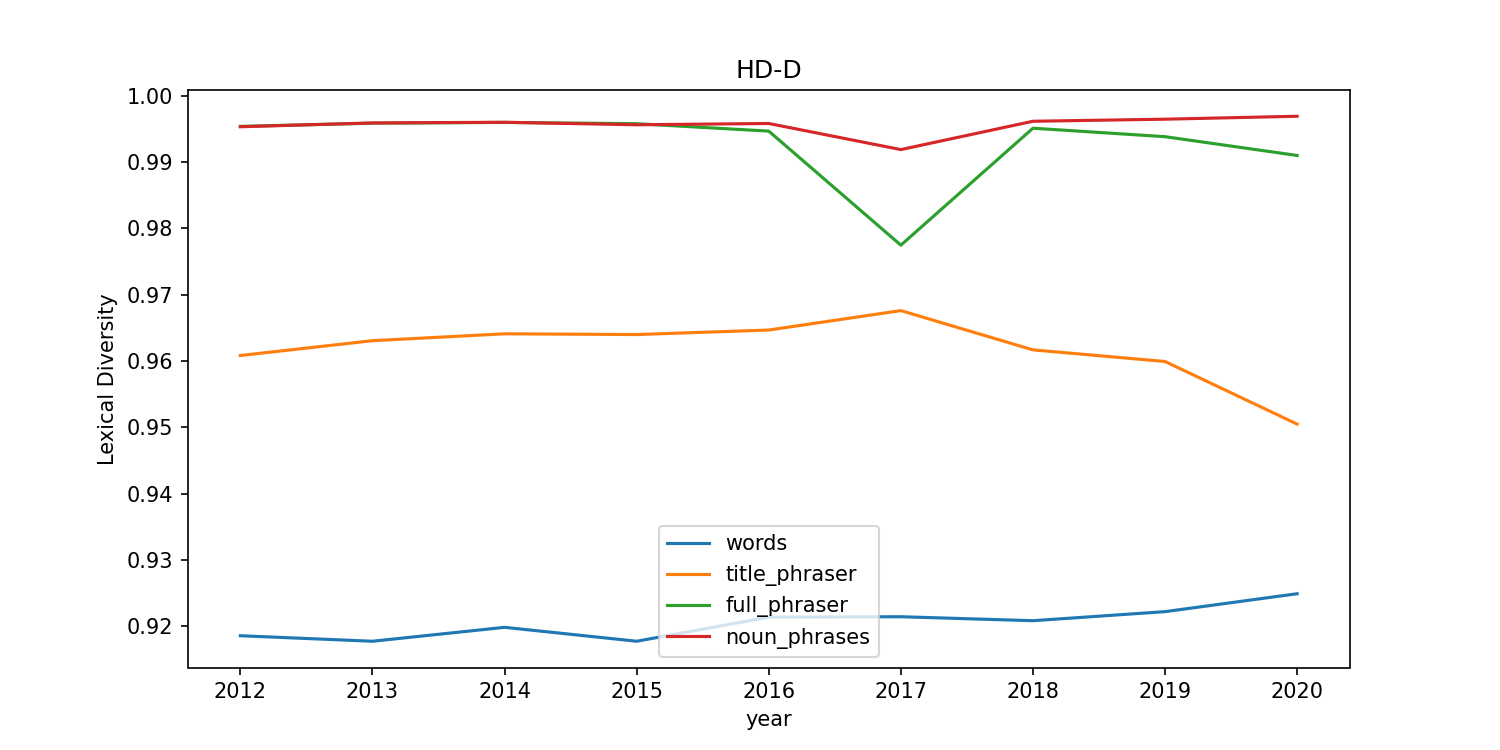}
  \caption{HD-D scores for for corpora variants over time}
    \label{fig:hdd}
\end{figure}
\begin{figure}[h]
  \centering
  \includegraphics[width=\linewidth]{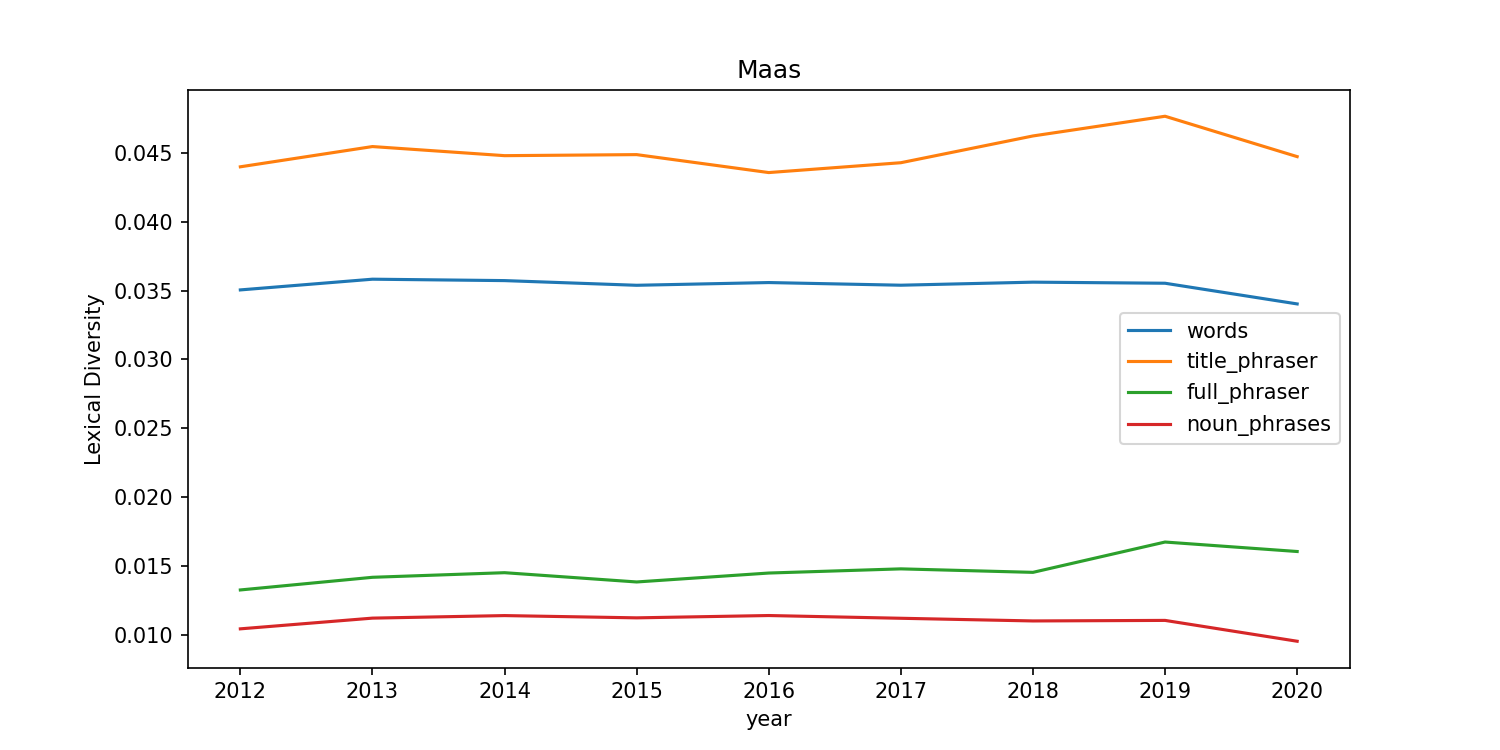}
  \caption{Maas index scores for for corpora variants over time}
  \label{fig:maas}
\end{figure}
\begin{figure}[h]
  \centering
  \includegraphics[width=\linewidth]{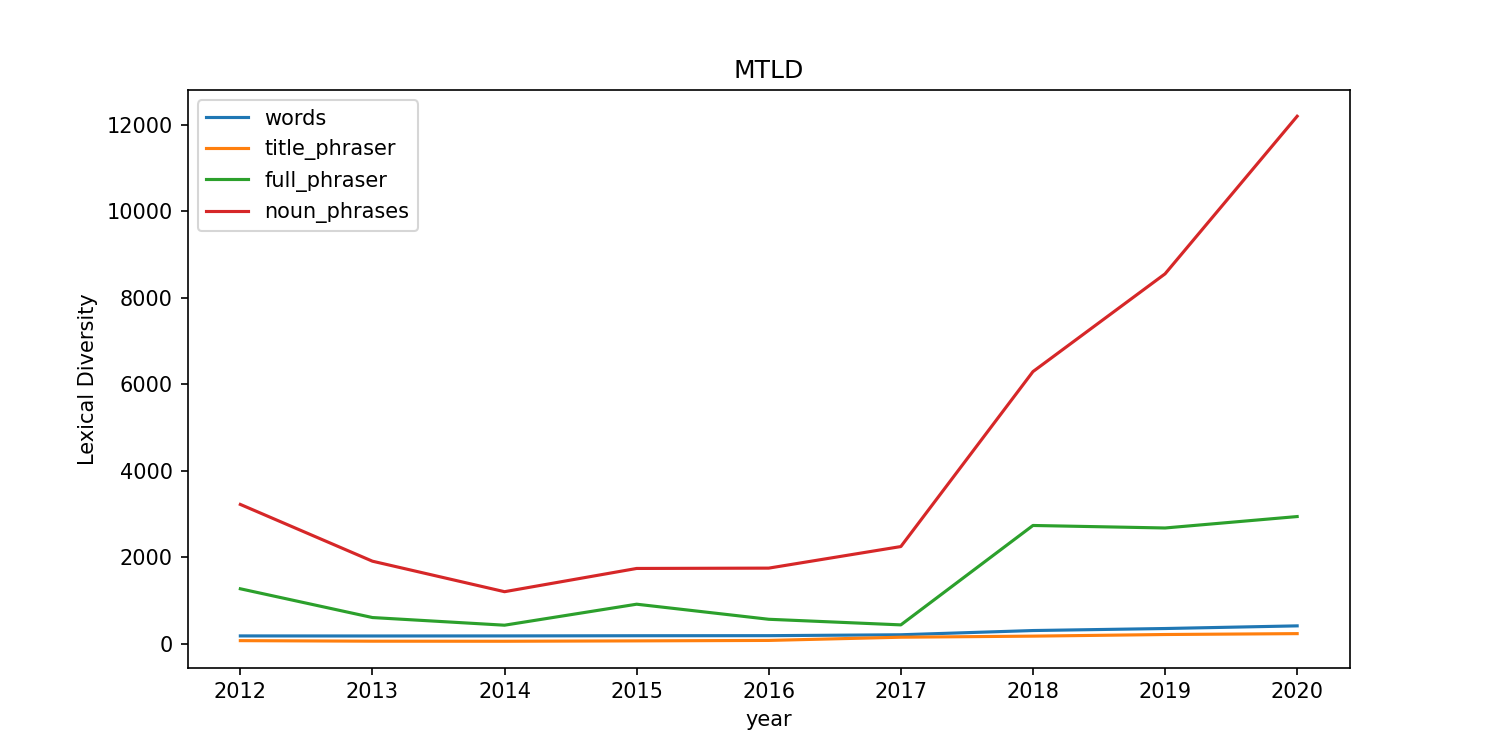}
  \caption{MTLD scores for for corpora variants over time}
 \label{fig:mtld}
\end{figure}
Compared to MTLD, the Maas index ~\ref{fig:maas} exhibited dramatically different results, with a relatively stable score across all years of the corpus, except for a drop at the end which is likely due to incomplete data for 2020. This drop provides some evidence that MAAS is more susceptible to text length than MTLD, which has been reported by others.

HD-D scores (figure ~\ref{fig:hdd}) for 2012-2020 exhibited relative stability until 2017 when the score for all words and for full-text phrases decreased, and the score for title phraser increased.  Table ~\ref{tabcorr} contains correlation scores between corpus types, by  lexical diversity time series scores.

Mindful of the potential impact of text length on the results, we reviewed the output statistics for raw text and derivative versions of the text corpus. It appears that there was a sustained increase in publication output starting after 2017 (table ~\ref{tabsummary} and in some metrics a commiserate shift in lexical diversity. Output increased by several over 1700 publications in 2017 and continued to increase in the three subsequent years. But as noted above, upwards trends in lexical diversity did not start until after 2017. One possibility is that there were a higher than usual number of institutional reports or whitepapers added in 2017.

Word counts per year (table ~\ref{tabsummary}) may also provide clues about lexical diversity scores. In 2016 there was a small dip in the number of unique words per title set per year, followed by an increase of almost 2000 words after 2017. In subsequent years, the unique word count remained almost constant. So it would appear that while MTLD did detect a shift in lexical diversity after 2017, the increase it reported for seemed out sized compared to the unique word count increase, but tracks more closely with the increase in the number of reports released. The HD-D index most accurately reflected the changes in unique word count by year. It also seemed not to be affected by the number of reports released per year. So it appears that for tracking year over year shifts in the lexical diversity of (our organization) report titles, the HD-D index performance seemed most likely to reflect an actual trend in the data with respect to new language and possibly indicative of new or expanding research, which manifested as a gradual upward trend until 2016 when a dip in lexical diversity occurred, followed by a more significant upward trend starting after 2017. This pattern seems to be real, confirmed by both the MTLD index scores and the HD-D index scores across multiple versions of the corpus.

The TTR and HD-D lexical diversity metrics were in close agreement over time: the noun phrases titles had the highest lexical diversity, followed by full text phrases, title phrases, with the original titles tokenized by word exhibiting the lowest scores. MTLD and Maas differed slightly: again noun phrases scored highest over time, with full text phrases, all words, and title phrases reported to have the lowest lexical diversity per these metrics. 

A number of studies have suggested that MTLD is the least sensitive to factors such as variable text length. This, combined with the fact that the scores reported over time by MTLD for all corpora variants were highly correlated lead us to conclude that this metric provides the most accurate representation of lexical diversity in our title corpora. 

\begin{table}[h]
\caption{Correlations among keyphrase models, lexical diversity scores 2012-2020}
\label{tabcorr}
\begin{tabular}{|l|l|l|l|l|}
\hline
\textbf{Sentence Variant} &\textbf{TTR}&\textbf{MTLD}&\textbf{Maas}&\textbf{HD-D}\\
\hline
All words vs Title phraser&\textbf{0.88}&\textbf{0.95}&0.23&-0.6\\
\hline
All words vs Full-text phraser&0.73&\textbf{0.92}&-0.3&-0.38\\
\hline
All words vs Noun phrases&\textbf{0.88}&\textbf{0.98}&\textbf{0.95}&0.13\\
\hline
\hline
Title phraser vs Full-text phraser&\textbf{0.88}&\textbf{0.85}&0.6&-0.22\\
\hline
Title phraser vs Noun phrases&0.57&\textbf{0.91}&0.069&-0.64\\
\hline
Full-text phraser vs Noun phrases&0.39&\textbf{0.93}&-0.29&\textbf{0.85}\\
\hline
\end{tabular}
\end{table}

\subsection{Analysis of the keyphrase sets}
Next we examine the discrete sets of phrased identified by each phrase extraction approach. We are particularly interested in the intersection of these sets, as well as gaining a better understanding of where they were not in agreement and whether these differences were in some way useful for improving the task of evaluating concept expansion. For example, we wanted to know if phrases identified by some of the more liberal models but not included in the smaller models were less interesting phrases, or vice-versa.

\begin{figure}[h]
  \centering
  \includegraphics[width=\linewidth]{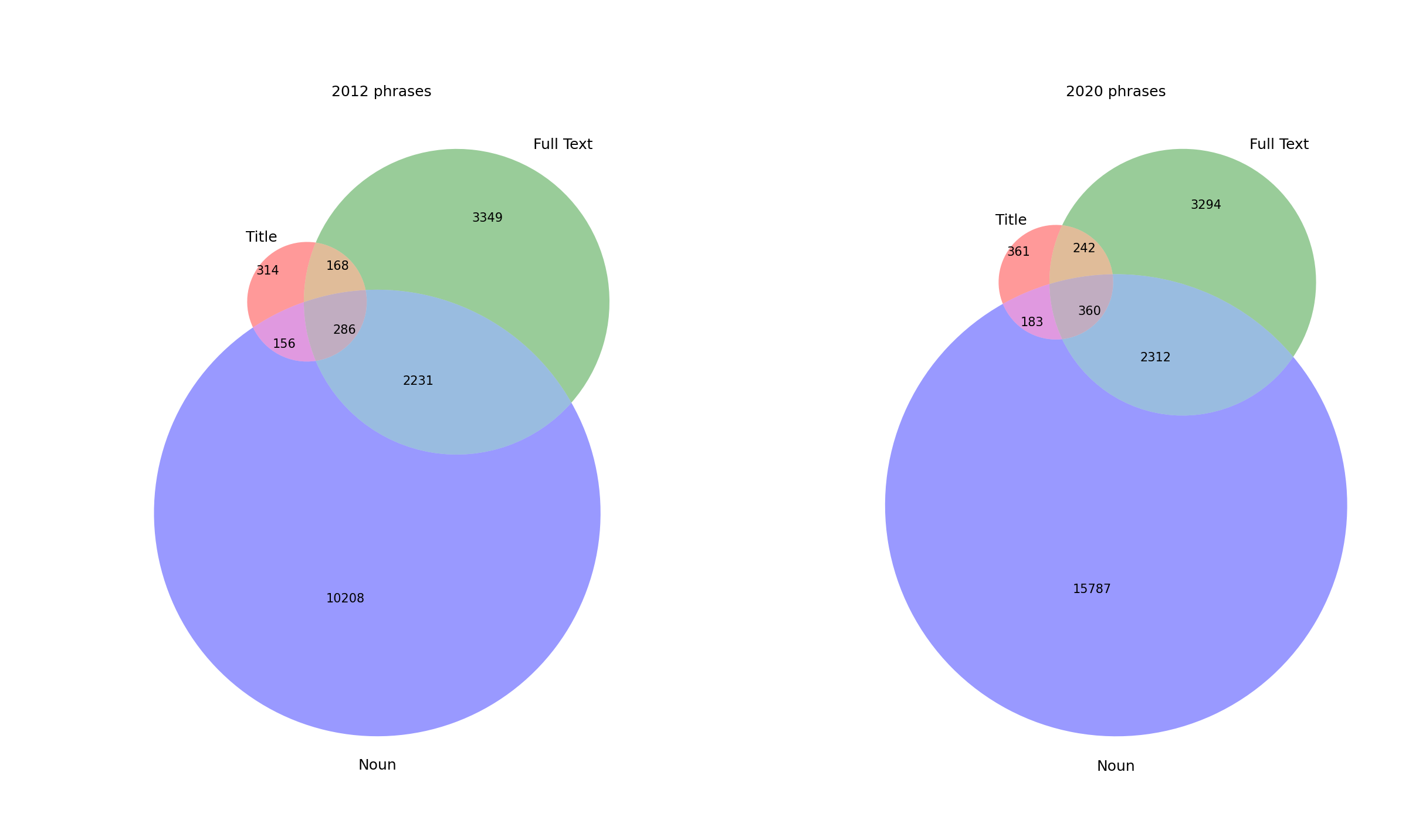}
  \caption{VENN diagram for keyphrase sets 2012 and 2020}
  \label{fig:venncombined}
\end{figure}

We found that there was a consistent ratio for the intersections between the corpus produced using the title phraser, full-text phraser, and the noun phraser. The noun phraser model always detected a larger number of phrases than the other two (figure 7), while the title phraser, trained on only the aggregated title corpus for 2012-2020, identified the smallest number of phrases. The size of each phrase collections also seem to be consistent in relation to one another from year to year (see Venn diagrams, figure ~\ref{fig:venncombined} ). But it was also apparent that each model yielded distinctly meaningful phrases that were not detected by the other models. In some cases, it was a matter of finer granularity, which perhaps counter-intuitively is represented by longer phrases.

Tables ~\ref{intersection} and ~\ref{differences} provide some specific examples of how the phrase models intersect in 2020, as well as how they differ. The full-text phraser was better at identifying named entities as well as important concepts. However some of the named entities were rather generic.  While they are clearly valid phrases, they obviously do not represent concepts or research topics. It would be interesting to specifically apply named entity detection over the titles to see if there is an intersection between full text phrases, noun phrases, and named entities that might form the basis of a "stop phrase" list.  It may make sense to eliminate phrases high frequency phrases that were detected by the full-text phraser, but not by the title phraser, perhaps by empirically establishing a frequency threshold value for this task. This may allow for automatic generation of stop phrases which could be safely removed by the pre-processing pipeline without requiring humans to produce or revise such a list, which was noted as a problem by \cite{milojevic2015quantifying}.

\begin{table*}[h]
\caption{Intersections of keyphrase sets for 2020, top 10 entries, by pair}
\label{intersection}
\centering
\begin{adjustbox}{width=1\textwidth}
\small
\begin{tabular}{|lc|lc|lc|}
\hline
\multicolumn{2}{|l|}{$\in$ Title phraser $\in$ Full-text phraser} & \multicolumn{2}{l|}{$\in$ Title phraser $\in$ Noun phrases} & \multicolumn{2}{l|}{$\in$ Full-text $\in$ Noun phrases}\\
\hline
machine learning&167&machine learning&27&national laboratory&31\\
uncertainty quantification&38&magnetic reconnection&13&machine learning&27\\
magnetic reconnection&25&wuantum annealing&13&national ignition facility&27\\
additive manufacturing&24&high explosives&11&magnetic reconnection&13\\
inertial confinement fusion&20&proton radiography&11&quantum annealing&13\\
quantum annealing&20&additive manufacturing&11&npdes outfall&12\\
deep learning&19&protoplanetary disks&11&high explosives&11\\
stainless steel&19&nuclear data&11&additive manufacturing&11\\
next generation&18&machine learning techniques&10&protoplanetary disks&11\\
high explosive&18&uncertainty quantification&10&proton radiography&11\\
\hline
\end{tabular}
\end{adjustbox}
\end{table*}

\begin{table*}[h]
\caption{Differences between keyphrase sets for 2020, top 10 entries}
\label{differences}
\centering
\begin{adjustbox}{width=1\textwidth}
\small
\begin{tabular}{|lc|lc|lc|}
\hline
\multicolumn{2}{|l|}{$\in$ Title phraser $\not\in$ Full-text phraser} & \multicolumn{2}{l|}{$\in$ Title phraser $\not\in$ Noun phrases} & \multicolumn{2}{l|}{$\in$ Full-text $\not\in$ Noun phrases}\\
\hline
using machine learning&46&using machine learning&46&national laboratory&31\\
nuclear data&33&high explosive&25&machine learning&27\\
high temperature&29&fuel cell&23&national ignition facility&27\\
earth system&28&equation of state&18&magnetic reconnection&13\\
case study&25&molecular dynamics&18&quantum annealing&13\\
training video&20&neutron multiplicity&15&npdes outfall&12\\
nuclear fuel&20&heavy flavor&14&high explosives&11\\
nuclear material&17&spent fuel&14&additive manufacturing&11\\
high performance&16&double shell&13&protoplanetary disks&11\\
density functional theory&15&ice sheet&13&proton radiography&11\\
\hline
\end{tabular}
\end{adjustbox}
\end{table*}

\section{Conclusion}
We identified and tested several approaches for detecting and extracting keyphrases from a text corpus. A corpus consisting of document titles poses some special problems for keyphrase detection, largely due to the relatively small number of examples of any given phrase that might be present in the data set. We found that some keyphrase detection models produce higher quality keyphrases than others, but all had some distinct benefits.  The use of a model trained on a larger but related corpus vastly improves the accuracy of keyphrase detection, because such a model has seen many more examples of potential keyphrases during training. Noun phrase detection based on part of speech tagging also identifies more keyphrases, but it is prone to identifying more common word pairings than other approaches such as PMI, such as article+noun pairings. Thus noun phrase detection is best used as a compliment to PMI, for example by evaluating the output from multiple approaches to identify the intersecting set of keyphrases. This intersection can be leveraged to refine a corpus so that it is more focused on concepts than extraneous information. Keyphrases found in one set but not another can be used to identify common phrases that may constitute the “institutional vocabulary” in a corpus. These can be used to produce a "keyphrase stop list" that can be used to further improve the input text.

We believe that the use of lexical diversity as a means to quantify ideas found in a document corpus is overall, problematic. As researchers such as McCarthy suggest, different metrics measure different characteristics of text segments. But these differences are ill-defined, and some metrics are prone to producing inconsistent results even when applied to the same corpus. Furthermore, there is no way to take into account synonymy or adjust for the presence of semantically related words or phrases. By reducing the corpus to keyphrases, we thought that a denser version of the input text converted to n-gram phrases might yield more consistent results, but with the exception of the MTLD metric, this appears not to be the case. The consistent correlated results produced by MTLD suggest that it does provide some coarse grained insight into the diversity of ideas expressed in a corpus over time. The inconsistent results of HD-D and Maas suggest that these metrics are not useful for this task. 

Jarvis suggested in 2013 \cite{jarvis2013capturing} that perhaps the whole notion of lexical diversity as a useful metric needs to be reconsidered, and in regard to quantifying the cognitive extent of publication titles, we concur. He pointed out that challenges similar to text understanding exist in other fields such as ecology, where there is a need to identify both the diversity of and dependency among actors in an ecosystem. He suggested a more holistic approach that treats a corpus as a complex system  might lead to better ways to characterize text, by considering qualities such as “actual number of types, effective types, evenness, disparity, importance, and dispersion.”  

Our contribution to this systems-oriented vision for text analysis has been to demonstrate that keyphrase detection can play an important role in evaluating the cognitive extent of a corpus and how it changes over time. Keyphrase detection is, we would argue, critical to this task. But there is still much room for improvement. Many of the qualities identified by Jarvis require that other NLP technologies such as word embeddings, ontologies, and language models be brought to bear on this problem. Otherwise it will be unclear exactly what it is we are measuring. 

 \bibliographystyle{elsarticle-num} 
 \bibliography{lexdiv}




\end{document}